\documentclass[twocolumn]{article}
\usepackage[utf8]{inputenc} 
\usepackage[T1]{fontenc}    
\usepackage{authblk}
\usepackage{hyperref}       
\usepackage{url}            
\usepackage{microtype}      
\usepackage{booktabs}       
\usepackage{multirow}
\usepackage{multicol}
\usepackage{amsmath}
\usepackage{amsfonts}
\usepackage{mathptmx}
\usepackage{nicefrac}
\usepackage{algorithm}
\usepackage{algorithmicx}
\usepackage{algpseudocode}
\usepackage[inline]{enumitem}
\usepackage{graphicx}
\usepackage{svg}
\usepackage{caption}
\usepackage{subcaption}
\usepackage{lipsum}

\usepackage{tikz} 
\usetikzlibrary{matrix,arrows,decorations.pathmorphing,shapes,calc,backgrounds,fit}
\tikzset{node style ge/.style={rectangle}}
\tikzset{>=latex}
\usepackage{environ} 
\usepackage{float} 

\makeatletter
\newsavebox{\measure@tikzpicture}
\NewEnviron{scaletikzpicturetowidth}[1]{%
  \def\tikz@width{#1}%
  \begin{lrbox}{\measure@tikzpicture}%
  \BODY
  \end{lrbox}%
  \pgfmathparse{#1/\wd\measure@tikzpicture}%
  \BODY
}
\makeatother

\definecolor{cbdarkblue}{HTML}{0173b2}
\definecolor{cborange}{HTML}{de8f05}
\definecolor{cbgreen}{HTML}{029e73}
\definecolor{cbred}{HTML}{d55e00}
\definecolor{cbpurple}{HTML}{cc78bc}
\definecolor{cbbrown}{HTML}{ca9161}
\definecolor{cbpink}{HTML}{fbafe4}
\definecolor{cbgray}{HTML}{949494}
\definecolor{cbyellow}{HTML}{ece133}
\definecolor{cblightblue}{HTML}{56b4e9}
\definecolor{cbdarkyellow}{HTML}{d7be0f}

\title{Multi-Omic Data Integration and Feature Selection for Survival-based Patient Stratification via Supervised Concrete Autoencoders\thanks{This preprint has not undergone peer review (when applicable) or any post-submission improvements or corrections. The Version of Record of this contribution was Accepted for publication on The 8th International Conference on machine Learning, Optimization and Data science - LOD 2022, please refer to that publication for the final version}}

\author[1,2,*,id]{\href{https://orcid.org/0000-0002-0347-7002}{Pedro~Henrique da~Costa~Avelar}}
\author[1,3,id]{\href{https://orcid.org/0000-0002-0118-4548}{Roman Laddach}}
\author[3,4,id]{\href{https://orcid.org/0000-0002-4100-7810}{Sophia N~Karagiannis}}
\author[2]{\href{https://orcid.org/0000-0003-0977-3600}{Min Wu}}
\author[1,*]{\href{https://orcid.org/0000-0001-8403-1282}{Sophia Tsoka}}

\affil[1]{Department of Informatics, Faculty of Natural, Mathematical and Engineering Sciences, King’s College London, United Kingdom}
\affil[2]{Institute for Infocomm Research, A*STAR, Singapore}
\affil[3]{St. John’s Institute of Dermatology, School of Basic \& Medical Biosciences, King’s College London, \& NIHR Biomedical Research Centre at Guy’s and St. Thomas’ Hospitals and King’s College London, Guy’s Hospital, King’s College London, United Kingdom}
\affil[4]{Breast Cancer Now Research Unit, School of Cancer \& Pharmaceutical Sciences, King’s College London, Guy’s Cancer Centre, United Kingdom}
\affil[*]{Corresponding authors \{pedro\_henrique.da\_costa\_avelar,sophia.tsoka\}@kcl.ac.uk}
\affil[id]{Author's name is a link to author's Orcid profile}

\date{Apr 2022}

\begin{document}
\flushbottom
\maketitle

\begin{abstract}
Cancer is a complex disease with significant social and economic impact. Advancements in high-throughput molecular assays and the reduced cost for performing high-quality multi-omics measurements have fuelled insights through machine learning . Previous studies have shown promise on using multiple omic layers to predict survival and stratify cancer patients. In this paper, we developed a Supervised Autoencoder (SAE) model for survival-based multi-omic integration which improves upon previous work, and report a Concrete Supervised Autoencoder model (CSAE), which uses feature selection to jointly reconstruct the input features as well as predict survival. Our experiments show that our models outperform or are on par with some of the most commonly used baselines, while either providing a better survival separation (SAE) or being more interpretable (CSAE). We also perform a feature selection stability analysis on our models and notice that there is a power-law relationship with features which are commonly associated with survival. The code for this project is available at: \url{https://github.com/phcavelar/coxae}
\end{abstract}

\section{Introduction}

Given the rapid advance of high-throughput molecular assays, the reduction in cost for performing such experiments and the joint efforts by the community in producing high-quality datasets with multi-omics measurements available, the integration of these multiple omics layers has become a major focus for precision medicine \cite{nicora2020integrated,bode2017precision}. Such integration involves analysis of clinical data across multiple omics layers for each patient, providing a holistic view of underlying mechanisms during development of disease or patient response to treatment.

Methods for multi-omic data integration can be classified in sequential, late, and joint integration approaches, depending on the order of implemented tasks and at what point multi-omics data is integrated \cite{uyar2021multi}. 
In sequential integration, each omic layer is analysed in sequence, i.e. one after the other, which was the used by many early approaches.
In late integration methods \cite{poirion2021deepprog,tong_deep_2020,wissel_hierarchical_2022,wang_similarity_2014,cabassi_multiple_2020}, each layer is analysed separately and then results are integrated, which helps capture patterns that are reproducible between different omics, but making the method blind to cross-modal patterns.
Finally, in joint integration methods \cite{chaudhary2018deep,zhang2018deep,ronen2019evaluation,asada2020uncovering,lee2020incorporating,uyar2021multi,wissel_hierarchical_2022, chalise_integrative_2017,shen_integrative_2009,meng_multivariate_2014,argelaguet_multiomics_2018,oconnell_rjive_2016}
all omics layers are analysed jointly from the start, with these methods often employing a dimensionality reduction method that maps all layers into a joint latent space representing all the layers \cite{cantini2021benchmarking}, making it possible to analyse cross-modal patterns that may help identify how multiple layers function and interact to affect a biological process.

All of the aforementioned approaches can be linked to particular challenges in the machine learning task. First, many of the datasets are sparse, often with some omics features missing between samples or studies, or even omics layers being unavailable for some patients. Second, molecular assay data are often highly complex, comprising thousands to tens-of-thousands of different features for each omics layer. Third, even with reduced cost of profiling, availability of data may still be prohibitively expensive and specialised, limiting the number of publicly-available datasets for analysis. Fourth, although experiments might be performed using the same assaying technologies and be collected with the same system in mind, there can be varying experimental conditions between datasets \cite{korsunsky2019fast} and batch effects may be present in the same dataset \cite{koch2018beginner}, which must be taken into account when analysing data, especially when analysis is attempted across multiple datasets.  Finally, recent advances have allowed profiling on the level of single cells, which increases dataset size dramatically, posing challenges to available methodologies.

In this context there are many end-tasks for which we might use multi-omic datasets: (Stratification) we might try to analyse which patients group together and analyse their clinical information to see what emerging patters are visible, including whether a patient is considered high or a low risk, whether a patient expression patterns are distinct from others, etc; (Classification) or we might even try to stratify patients into previously known groups, such as whether a patient is in a previously-defined subtype group, whether they might respond or not to a certain drug or treatment, etc; (Regression) And we can also try to infer directly to which level a patient might react to something, such as reaction to a treatment or drug, or how long is a patient is expected to survive given his conditions, etc. (Biomarker identification) given any of the aforementioned tasks, we might also try to interpret why are patients stratified or classified in a certain way, or why they will react to the level predicted by the algorithm, this generally entails in identifying which biomarkers (that is, which features) are associated with these responses. 

The benefits of integrated datasets include more accurate patient stratification (e.g. high/low risk), disease classification or prediction of disease progression. All of those may suggest better treatment strategies, resulting in better patient outcomes. Additionally, the combined information may also be used for biomarker identification supporting further research.

In this paper we provide several contributions to the field of survival-based autoencoder (AE) integration methods, which can be summarized in the following points:
\begin{enumerate*}
    \item In Subsection~\ref{ssec:coxsae}, we develop a simpler Supervised Autoencoder (SAE) as an alternative to the HierSAE model \cite{wissel_hierarchical_2022} for data integration, a method which provides stable and efficient survival separation, and use it as an upper-bound baseline for performance testing our concrete supervised-autoencoder.
    \item In Subsection~\ref{ssec:concretesae}, we propose the Concrete Supervised-Autoencoder (CSAE), building up on Concrete Autoencoders \cite{balin2019concrete}, a method for supervised feature selection, which we showcase with the case study of survival-based feature selection.
    \item In Subsection~\ref{ssec:pipeline}, we provide a testing framework more stringent than that used by previous work with which we compare our results with a standard PCA pipeline as well as the more advanced Maui \cite{ronen2019evaluation} method.
    \item With our testing framework, in Section~\ref{sec:results}, we show that the Concrete Supervised Autoencoder has achieved performance on par with that of more complex baselines, while simultaneously being more interpretable, and also provide, to the best of our knowledge, the first feature importance analysis with multiple runs on a model of such a family.
\end{enumerate*}

\section{Related Work}

\subsection{Autoencoders for Multi-Omics}

After an initial publication in 2018 showing the use of Autoencoders (AEs) for dimensionality reduction in multi-omic datasets \cite{chaudhary2018deep}, there has been a wave of re-application of this technique in cancer risk separation, prognostication, and biomarker identification \cite{chaudhary2018deep,zhang2018deep,asada2020uncovering,lee2020incorporating,poirion2021deepprog}. All of the applications of these methods share the same pipeline, and most \cite{chaudhary2018deep,zhang2018deep,asada2020uncovering,poirion2021deepprog} use the same techniques up to the AE optimisation, having only minor differences in the hyperparameters and loss functions. Only \cite{lee2020incorporating} has a major difference in their model, using Adam instead of SGD as the optimiser. Another publication in this vein is \cite{poirion2021deepprog}, which does not perform early-integration, instead opting to project the input for each omics layer separately, concatenating features from different omics layers after Cox-PH selection, and using a boosting approach to train and merge multiple models into a single predictor. One of the main differences from these methods and ours is that the AEs are used to perform risk subgroup separation on the whole dataset, which is then used as ground truth for another classification model, whereas our pipeline is entirely cross-validated.

In \cite{ronen2019evaluation} a method was proposed using a Variational Autoencoder (VAE), dubbed Maui, to learn reduced-dimensionality fingerprints of multiple omics layers for colorectal cancer types, showing that their method both correctly mapped most samples into the existing subtypes, but also identified more nuanced subtypes through their approach, while still keeping a level of interpretability by relating input features with embedding features through correlation. The same group expanded their analysis on a pan-cancer study \cite{uyar2021multi}, changing their interpretability approach to consider the absolute value of the multiplication of the neural path weights for each input-fingerprint feature pair. This interpretability is one of the many methodological differences that sets these works apart from the aforementioned AE approaches based on \cite{chaudhary2018deep}. These VAE-based works also use the fingerprints to cluster samples into different risk subgroups and for hazard regression. The main difference with our proposed framework is the type of AE used (VAE instead of AE), and that our models are supervised with a Cox loss and that our Concrete Supervised Autoencoder uses a different type of encoding function.

\subsection{Supervised Autoencoders}

One can also perform Cox regression on neural networks \cite{katzman2018deepsurv,ching2018cox,huang2019salmon}, and this obviously implies that one can add a hazard-predicting neural network block on an Autoencoder's fingerprints. Independently from our Cox-SAE model, presented in Subsection~\ref{ssec:coxsae}, two other works developed similar techniques. In \cite{tong_deep_2020}, they developed the same principles of performing Cox-PH regression on the fingerprints generated by the autoencoder. However, the main difference is that the integration is done on the fingerprint-level -- that is, they perform dimensionality reduction through the autoencoder as normal, and then either concatenate the fingerprints to perform Cox-PH regression
(Figure~4 of their paper)
, or they do cross-omics decoding, with the Cox-PH loss being calculated on the average of both generated fingerprints
(Figure~5 of their paper)
. They also limit themselves to 2-omics integration. A recent paper also improved on this idea by proposing an autoencoder which tries to reconstruct the concatenation of the fingerprints generated through the other encoders \cite{wissel_hierarchical_2022}
, as seen in Figure~1D of their paper,
while performing a 6-omics integration also using clinical data.

\subsection{Concrete Autoencoders}

Recently, the efficacy of using a concrete selection layer as the encoder of an autoencoder was shown \cite{balin2019concrete}, dubbing this model the Concrete Autoencoder, and providing tests with many different feature types, including gene expression for providing an alternative to the ``943 landmark genes'' \cite{lamb_connectivity_2006}, as well as mice protein expression levels. A concrete selection layer \cite{maddison_concrete_2017} is an end-to-end differentiable feature selection method, that uses the reparametrisation trick \cite{kingma_auto-encoding_2014}, to provide a continuous approximation of discrete random variables, which Balin et al. used in its autoencoder model with an exponentially decreasing temperature during training to provide a smooth transition from random feature selection to discrete feature selection \cite{balin2019concrete}.

\section{Methods}

\subsection{Datasets}

We wanted to test our models on open high-quality cancer data with multiple omics layers and which had survival information. The TCGA datasets fit these criteria, being used as a baseline testing dataset for many developed methods, including the most relevant related work \cite{chaudhary2018deep,ronen2019evaluation,uyar2021multi,poirion2021deepprog,tong_deep_2020,wissel_hierarchical_2022}. We use the datasets provided by \cite{wissel_hierarchical_2022} and described in Table~\ref{tab:coxae-dsets}, following the same preprocessing steps, however we use our own set of splits for cross validation, as we perform 10-fold cross validation, with 10 repeats, as compared to 2 repeats of 5-fold cross validation in the initial work. 

\begin{table*}
    \centering
    \begin{tabular}{lrrrrrrrrrrr}
        \toprule
        Dataset & Samples & Clinical & GEx & CNV & Methylation & $\mu$RNA & Mutation & RPPA & Total & Used\\
        \midrule
        BLCA & 325 & 9 & 20225 & 24776 & 22124 & 740 & 16317 & 189 & 84380 & 4938 \\
        BRCA & 765 & 9 & 20227 & 24776 & 19371 & 737 & 15358 & 190 & 80668 & 4936 \\
        COAD & 284 & 16 & 17507 & 24776 & 21424 & 740 & 17569 & 189 & 82221 & 4945 \\
        ESCA & 118 & 17 & 19076 & 24776 & 21941 & 737 & 9012 & 193 & 75752 & 4947 \\
        HNSC & 201 & 16 & 20169 & 24776 & 21647 & 735 & 11752 & 191 & 79286 & 4942 \\
        KIRC & 309 & 14 & 20230 & 24776 & 19456 & 735 & 9252 & 189 & 74652 & 4938 \\
        KIRP & 199 & 5 & 20178 & 24776 & 21921 & 738 & 8486 & 190 & 76294 & 4933 \\
        LGG & 395 & 15 & 20209 & 24776 & 21564 & 740 & 10760 & 190 & 78254 & 4945 \\
        LIHC & 157 & 3 & 20078 & 24776 & 21739 & 742 & 8719 & 190 & 76247 & 4935 \\
        LUAD & 338 & 11 & 20165 & 24776 & 21059 & 739 & 16060 & 189 & 82999 & 4939 \\
        LUSC & 280 & 20 & 20232 & 24776 & 20659 & 739 & 15510 & 189 & 82125 & 4948 \\
        OV & 161 & 17 & 19064 & 24776 & 19639 & 731 & 8347 & 189 & 72763 & 4937 \\
        PAAD & 100 & 26 & 19932 & 24776 & 21586 & 732 & 9412 & 190 & 76654 & 4948 \\
        SARC & 190 & 45 & 20206 & 24776 & 21724 & 739 & 8385 & 193 & 76068 & 4977 \\
        SKCM & 238 & 3 & 20179 & 24776 & 21635 & 741 & 17731 & 189 & 85254 & 4933 \\
        STAD & 304 & 7 & 16765 & 24776 & 21506 & 743 & 16870 & 193 & 80860 & 4943 \\
        UCEC & 392 & 24 & 17507 & 24776 & 21692 & 743 & 19199 & 189 & 84130 & 4956 \\
        \bottomrule
    \end{tabular}
    \caption{Number of features in each of the used TCGA datasets. The ``Used'' column indicates how many features we expect the models to use after the second pipeline step, which involved selecting the top 1000 features for each omics layer. All datasets were used as preprocessed and made available by \cite{wissel_hierarchical_2022}.}
    \label{tab:coxae-dsets}
\end{table*}

Some points raised in the literature about these datasets are interesting to be reiterated here: The LUSC and PRAD datasets were considered to be some of the hardest ``(...) As the default, we use 10 models with 80\% of original training samples to construct all the cancer models, except for LUSC and PRAD which we use 20 models since they are more difficult to train. (...)'' \cite{poirion2021deepprog}; The combination of Clinical factors and Gene Expression is said to perform better than using multiple omics layers with regards to performance on simpler models  \cite{wissel_hierarchical_2022}.

\subsection{Evaluation and Metrics}

\subsubsection{Concordance Index}

The most commonly-used quantitative metric for both Survival Regression and and Survival Stratification is the Concordance-Index (C-Index), which can be seen as a generalisation of the AUC metric for regression, being similarly interpreted, with a C-Index of 0 representing perfect anti-concordance, 1 representing perfect concordance, and 0.5 being the expected result from random predictions. The metric is calculated by analysing the number of times a set of model predictions $f(x_i) > f(x_j)$ given that $y_i > y_j$ as well, while also handling censored data, due to the fact that if a value $y_j$ is censored, it is less certain to say that $y_i$ is in fact greater than $y_j$. That is, given a set of features $X$ to which a function $f$ is applied to, and the ground-truth consisting of both the set event occurrences $E$ as well as the drop-out times $Y$, we would have the metric defined as:

\begin{equation*}
\operatorname{CI}(f(X),Y,E) =
    \frac{
        \operatorname{CP}(f(X),Y,E) +
        \frac{
            \operatorname{TP}(f(X),Y,E)
        }{
            2
        }
    }{
        \operatorname{AP}(f(X),Y,E)
    }
\end{equation*}

Where $\operatorname{CI}(f(X),Y,E)$ if the concordance index, $\operatorname{CP}(f(X),Y,E)$ is the number of correct pairs, $\operatorname{TP}(f(X),Y,E)$ the number of tied pairs, and $\operatorname{AP}(f(X),Y,E)$ the number of admissible pairs. An admissible pair is one that both events were observed or where a single event $e_i$ was observed and $y_i \leq y_j$. The number of correct and tied pairs are be taken from the only from the admissible pairs. To the best of our knowledge, none of the related work has the entire pipeline validated as we show here, with most related work normalising the whole dataset before the pipeline \cite{chaudhary2018deep,ronen2019evaluation,tong_deep_2020,wissel_hierarchical_2022}

\subsubsection{Qualitative Analysis}

We can also perform a qualitative analysis of the models by analysing how well the expected survival of subgroups classifies when using the method as an analysis method. One way to qualitatively assess a model for the Stratification task would be to fit Kaplan-Meier (KM) curves for each subgroup the models stratifies, and then analysing the behaviour of each subgroup. This has been done in many of the related works, where they report the KM curves for all the samples, accompanied of the logrank p values for the subgroup separation \cite{chaudhary2018deep,ronen2019evaluation,uyar2021multi,poirion2021deepprog}.

\subsection{Main Testing Pipeline} \label{ssec:pipeline}

We followed the common and well-established practice of cross validation of the whole pipeline, During our preliminary testing, performing scaling only on training samples versus on the whole dataset accounted for a drastic performance change. Also, analysing logrank p-values and concordance indexes on the whole dataset generally means that the logrank p values will be significant only due to the fact that the training dataset is generally larger than the test dataset, skewing the results towards already-seen data. The testing framework we adopt solves both these issues, and this testing framework is one of the biggest methodological differences between our model and some previous work \cite{chaudhary2018deep,zhang2018deep,asada2020uncovering,lee2020incorporating,poirion2021deepprog,ronen2019evaluation,uyar2021multi}, which either provide external validation through other cohorts and/or validate piecewise.

\begin{figure*}
    \centering
    \includegraphics[width=.9\textwidth]{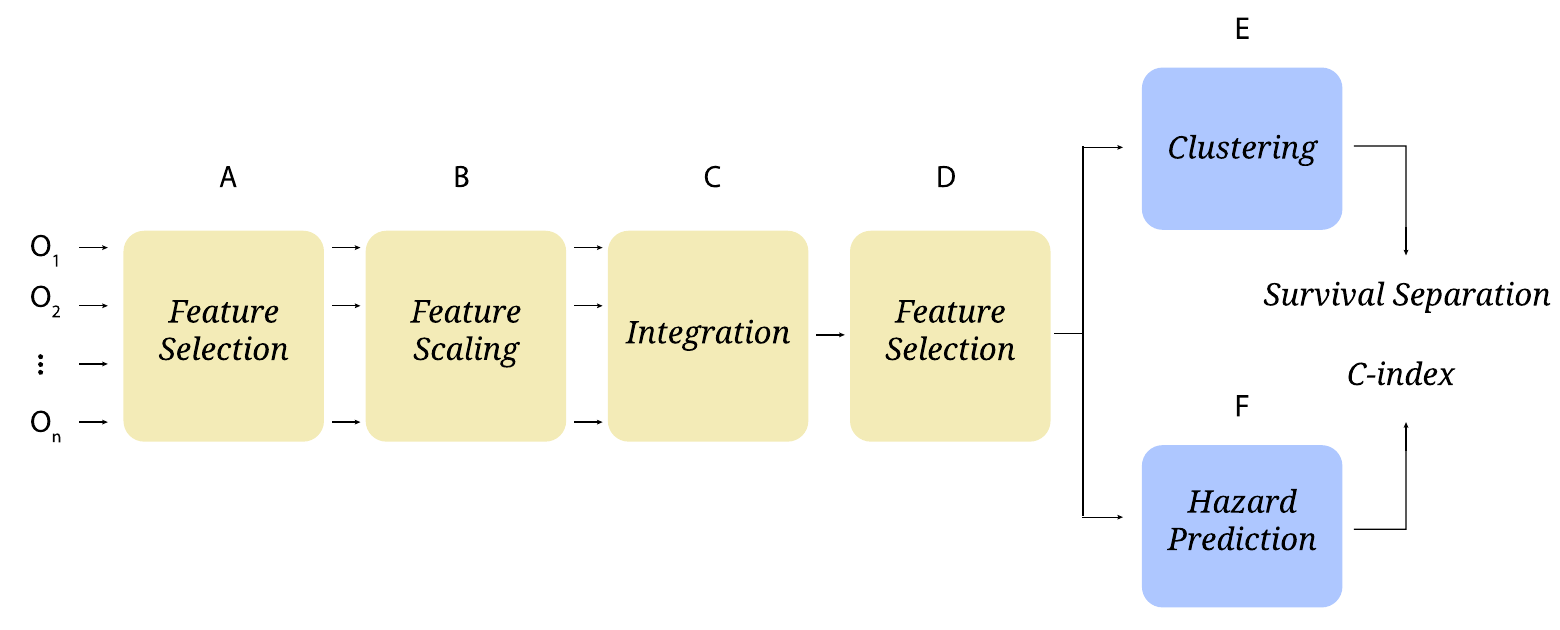}
    \caption{A diagram showing the pipeline used in our testing framework.}
    \label{fig:pipeline}
\end{figure*}

As part of our pipeline we perform 6 steps, shown with letters A to F in \ref{fig:pipeline}:
\begin{enumerate*}[label=(\Alph*)]
    \item First perform per-omics feature selection by selecting the $k$ most variable features from each omics layer, as was done in previous work \cite{ronen2019evaluation,tong_deep_2020,poirion2021deepprog,uyar2021multi};
    \item After that, doing feature scaling by z-scoring the selected features, here being different from most related work in that we only perform feature scaling with those samples available during training;
    \item Then, perform feature compression (i.e. integration, selection, or linear combination) through the model in question;
    \item After this, performing Cox-PH univariate feature selection to select those features that are considered to be relevant for survival, and then;
    \item Performing 2-clustering (with a clustering algorithm that allows for clustering new inputs) on this integrated dataset, which can be used for survival subgroup separation and differential expression analyses;
    \item Or performing hazard prediction, which can be used to rank patients with regards to survival probability, and is also the endpoint from which we can calculate C-indexes for the pipeline.
\end{enumerate*}

We run all of the experiments on a HPC cluster, creating a single process for each dataset, using the same consistent seeds across algorithm repetitions to ensure the same 10-fold cross validation splits were used for all models. For the BRCA and STAD we ran 5 repetitions of 10-fold cross validation due to limitations in our compute budget, and for all other datasets we used 10 repetitions of 10-fold cross validation. For the BRCA, STAD and KIRP datasets we ran the experiments with 16000MB of available RAM memory, and for all other datasets we made available 8000MB of RAM. All experiments were run with 16 cores available to the program.

All models use:
\ref{fig:pipeline}:
\begin{enumerate*}[label=(\Alph*)]
    \item $k\leq1000$ for feature selection on each omics layer, which gives us the ``Used'' column in Table~\ref{tab:coxae-dsets};
    \item We then proceed to perform feature scaling using the mean and standard deviation available in the training dataset for each fold;
    \item Then, we use as a default 128 target ``fingerprint'' features for all models, and on traditional autoencoder-based models we choose 512 neurons on the hidden layer to give us roughly 10x, and then a further 5x compression on the input feature size. The Maui model was trained for 400 epochs (taken as a default from their codebase) whereas our models were trained for 256 epochs, with the Adam optimiser using $0.01$ as a learning rate and $0.001$ as the l2 normalisation weight. All of the AE models implemented by us used $0.3$ dropout rate and a gaussian noise with zero mean and $0.2$ standard deviation added to input features during training, and we used rectified linear units as a nonlinearity on all intermediate layers. We used the same temperature settings provided in \cite{balin2019concrete} for our concrete selection layers, starting with a temperature of $10$ and ending with a temperature of $0.1$;
    \item Cox-PH univariate ``fingerprint'' feature selection is done with a significance threshold of $p<0.05$, falling back on using all the of the fingerprints if no fingerprint is identified as significant for survival;
    \item 2-clustering was done with KMeans with 10 initialisations, using the best in terms of inertia, with a maximum of 300 iterations and a tolerance of $0.001$.;
    \item All Cox-Regression was done non-penalised, unless the model failed to converge, in which case we did Cox-Regression with $0.1$ penalisation.
\end{enumerate*} Furthermore, if a model fails to run on any fold, we drop that value. Since we perform 10 repetitions of the 10-fold cross validation, this means that a model has at least some results for each dataset, but the PCA model failed to produce any results on two datasets due to convergence problems. 

\subsection{Cox-Supervised Autoencoder} \label{ssec:coxsae}

Many methods available in the literature have used, or attempted to use, autoencoders to perform dimensionality reduction and then select survival-relevant features from the autoencoder fingerprints through Cox-PH regression \cite{chaudhary2018deep,zhang2018deep,ronen2019evaluation,asada2020uncovering,lee2020incorporating,poirion2021deepprog,uyar2021multi}. We would like to argue that the hidden assumption contained within an autoencoder loss function is insufficient to provide features relevant for survival, due to the fact that feature combinations that are good at predicting other features might not necessarily be good for survival prediction, an argument that has support with preliminary tests where the non-trained models perform just as good as the trained models. To solve these issues in this section we propose our independently developed method of a Cox-Supervised Autoencoder (SAE) that addresses this issue.

\begin{figure}
    \centering
    \includegraphics[width=.5\textwidth]{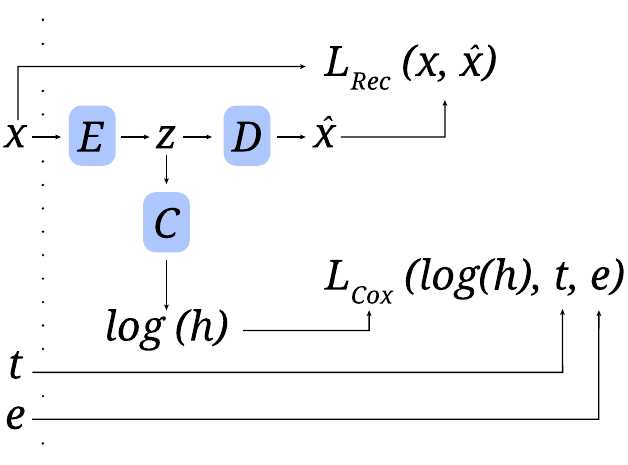}
    \caption{A diagram visualising how a Cox-Supervised Autoencoder (SAE) model works. We use both the multi-omics input $x$ to generate the encoding $z$ and the reconstruction $\hat{x}$, as well as calculate the log hazards $log(h)$ from $z$ using another neural module, which is trained using a Cox-PH loss alongside the time and event informations, $t$ and $e$.}
    \label{fig:coxae-diagram}
\end{figure}

To solve the lack of an inductive bias towards survival, we would like to introduce a Cox-PH model inside the neural network as a normalising loss, so that the model learns not only to create codes which are good at reconstructing the input, but also codes that are indicative of survival. Since a Cox-PH model is end-to-end differentiable, this is easily done by simply adding a Cox-PH model on the generated fingerprints, and performing Cox-PH regression with regards to the input survival times and observed events. We can see a schematic overview of such a model in Figure~\ref{fig:coxae-diagram}.

In mathematical terms, then, we would have a model composed of the three neural network blocks shown in Figure~\ref{fig:coxae-diagram}: An encoder $E$, which takes the input $x$ and produces a set of fingerprints $z$ from which a decoder $D$ produces a reconstruction of the input $\hat{x}$, and finally a linear layer $C$ which uses the same fingerprints $z$ to produce a log-hazard estimate $log(h)$ for each patient. We then perform gradient descent on a loss $L$ which is composed not only of a reconstruction loss $L_{Rec}(x,\hat{x})$ and a normalisation loss $L_{norm}(E,D,C)$ on the weights of $E$, $D$, and $C$, but also the Cox-PH regression loss $L_{Cox}(log(h),t,e)$ using the information available about the survival time $t$ and the binary information of whether the event was censored or not $e$, giving us the equation below:

\begin{equation}\label{eq:sae}
\begin{aligned}
    L(x,t,e,E,D,C) = &L_{rec}(x,D(E(x))) +
    \\&
    L_{cox}(C(E(x)),t,e) +
    \\&
    L_{norm}(E,D,C)
\end{aligned}
\end{equation}

Here we use the Cox-PH loss from \cite{katzman2018deepsurv} as implemented by the pycox library (\url{https://github.com/havakv/pycox}) version 0.2.3, which, assuming that $x$, $t$, and $e$ are sorted on $t$, and that we have $k$ examples, is defined as:

\begin{equation}
\begin{aligned}
    L_{cox}(log(h)),t,e) &= \frac{\sum_{1 \leq i \leq k} (log(h_i)) - log(g_i) + \gamma}{\sum_{1 \leq i \leq k} e_i}, \\
    g_i &= \sum_{1 \leq j \leq j} e^{log(h_j) - \gamma}, \\
    \gamma &= max(log(h))
\end{aligned}
\end{equation}

For the reconstruction, we chose the Mean Square Error (MSE) loss, due to its symmetry, as below:

\begin{equation}
    L_{rec}(x,\hat{x}) = \frac{\sum{1 \leq i \leq k}\sum{1 \leq j \leq d} (x_{i,j}-\hat{x_{i,j}})^2}{k}
\end{equation}

And, finally, we use the L2 norm of the model weights as our normalisation loss, using the Frobenius norm $||\cdot||_F$ and a hyperparameter $\lambda$ to control the how much of the norm is applied:

\begin{equation}
    L_{norm}(E,D,C) = \lambda \sum_{w \in E,D,C} ||w||^2_F
\end{equation}

Note that in our definitions here, the autoencoder receives as input all of the omics layers at the same time, much like many models in the literature \cite{chaudhary2018deep,ronen2019evaluation,uyar2021multi}, and we argue that this provides integration, since all of the fingerprints may be composed of combinations of features from different omics levels. This is highly different from models where each omics layer is used as an input to a separate autoencoder, and then concatenated \cite{poirion2021deepprog,tong_deep_2020}; or otherwise combined through pooling \cite{wissel_hierarchical_2022}; or even through an hierarchical autoencoder \cite{wissel_hierarchical_2022}, which only then would de-facto integrate the omics layers through the separately-compressed layer fingerprints through a much more complex procedure.

\subsection{Concrete Supervised Autoencoder} \label{ssec:concretesae}

\begin{figure*}
    \centering
    \begin{subfigure}[b]{0.4\textwidth}
        \centering
        \includegraphics[width=\textwidth]{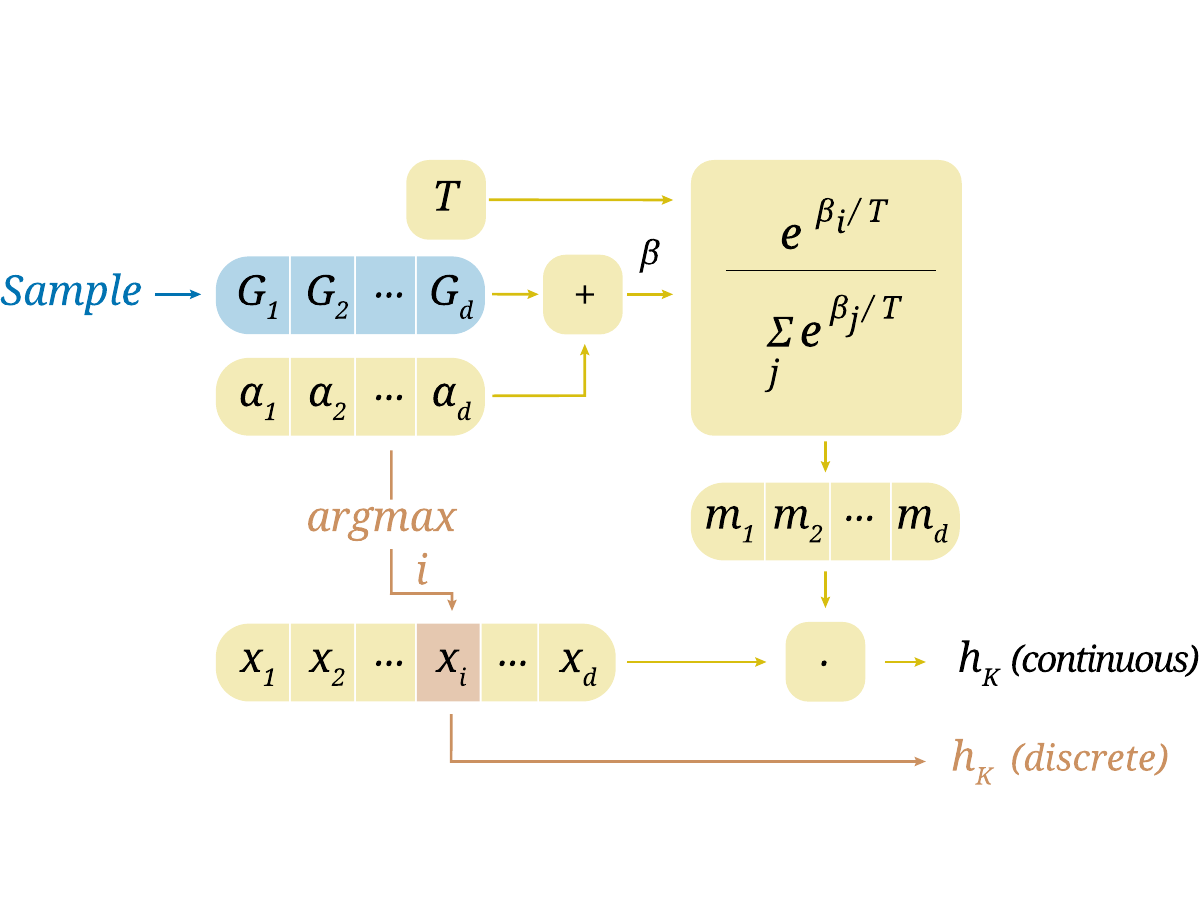}
        \caption{}
        \label{subfig:concrete-selection-layer}
    \end{subfigure}
    \begin{subfigure}[b]{0.4\textwidth}
        \centering
        \includegraphics[width=\textwidth]{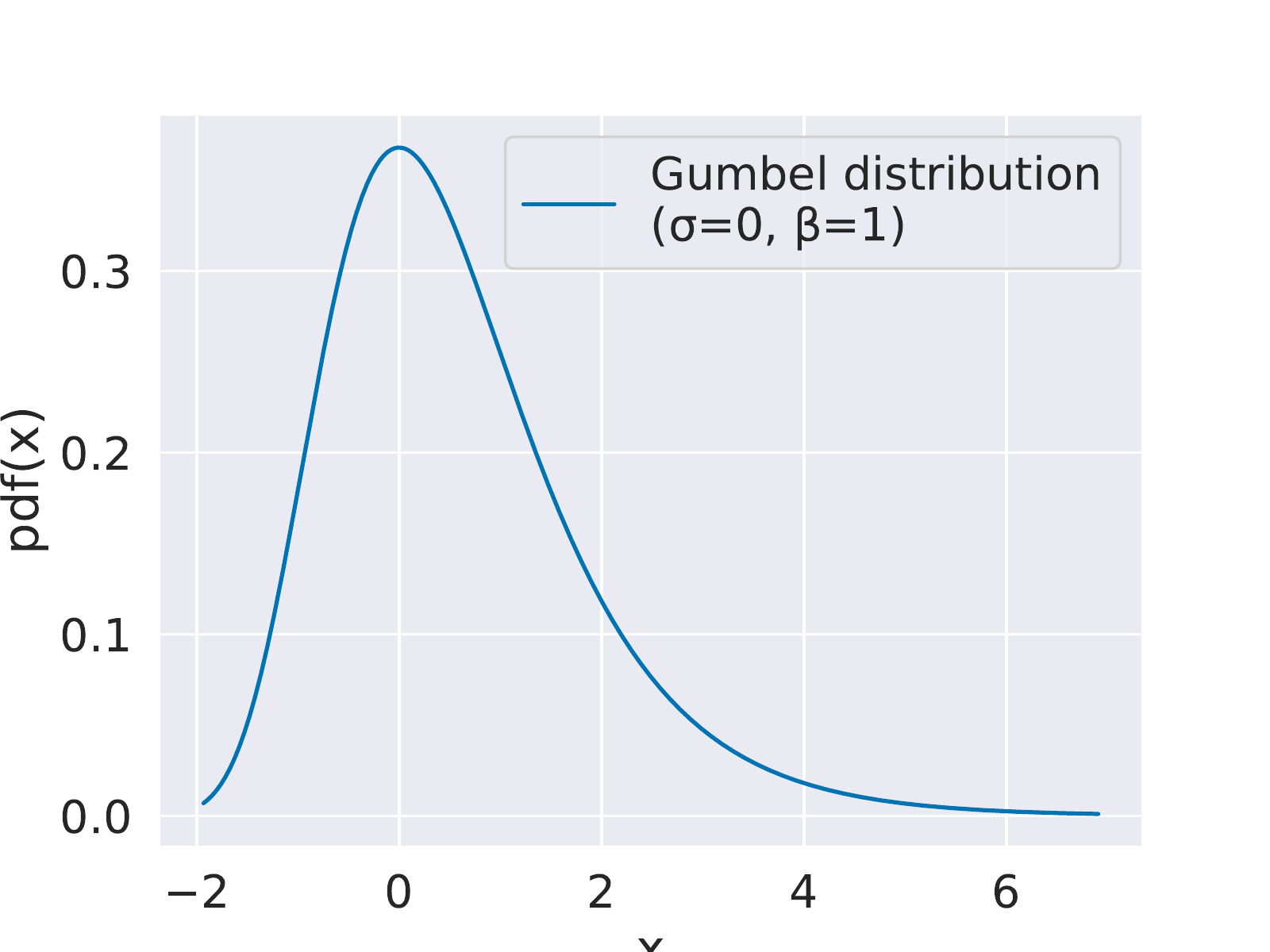}
        \caption{}
        \label{subfig:gumbel-distribution}
    \end{subfigure}
    \caption{A diagramatic Concrete Selection Layer's neuron (\subref{subfig:concrete-selection-layer}), where {\color{cbdarkblue}blue} elements represent continuous stochastic nodes, {\color{cbdarkyellow}yellow} elements are continuous and deterministic, and {\color{cbbrown}brown} elements show the discrete selection path, generally followed after training. The values of $G_i$ are sampled from the Gumbel distribution, whose PDF we can see on (\subref{subfig:gumbel-distribution}).}
    \label{fig:concrete-selection}
\end{figure*}

Another possible point of concern for many of the multi-omics analysis pipelines is that using neural-network-based models can lead to less interpretable results. To address this, we use the Concrete Autoencoder proposed by \cite{balin2019concrete} to build a Concrete Supervised Autoencoder (CSAE), using the same concrete selection layer and reparametrization tricks as described previously \cite{kingma_auto-encoding_2014,maddison_concrete_2017,balin2019concrete}. We can see a diagramatic representation of a concrete selection layer and the gumbel distribution, used in its training, in Figure~\ref{fig:concrete-selection}.

Thus, our encoder follows the same exponential decay rate, and model reparametrisation described previously \cite{balin2019concrete}, to give us a sampling in $d$ dimensions with parameters $\alpha \in \mathcal{R}^d, \alpha_i > 0 \forall \alpha_i$, regulated by a temperature $T(b)$, and with d values $g \in \mathcal{R}^d$ sampled from a Gumbel distribution, giving us a probability distribution $m \in \mathcal{R}^d$: 

\begin{equation}
g_j = \frac{e^{log(\alpha_j+g_j)/T(b)}}{\sum_{k=1}^{d} e^{log(\alpha_k+g_k)/T(b)}}
\end{equation}

During training, each element outputted by a concrete selection ``neuron'' $i$, would be a linear combination of the input $x \in \mathcal{R}^d$:

\begin{equation}
E(x)_i = x_i*g_j
\end{equation}

But as the temperature $T(b) \rightarrow 0$, we have that each node will select only a single input, at which point we can switch from the linear combination to simple indexing:

\begin{equation}
E(x)_i = \arg \max_j \alpha_j
\end{equation}

This allows us to smoothly transition for feature combination and also to reduce evaluation-time memory requirements, since the indexing takes $O(1)$ space per neuron instead of $O(d)$ of the linear approximation. To perform such smooth transition, we use the previously defined exponential temperature decay schedule, with $T_0$ being the initial temperature, $T_B$ the minimum temperature, $b$ being the current epoch, and $B$ the maximum number of epochs:

\begin{equation}
    T(b) = T_0(T_B/T_0)^{b/B}
\end{equation}

Having our encoder thus defined, we simply apply the same optimisation as in Equation~\ref{eq:sae}, replacing the traditional pyramidal MLP used as the encoder for the Autoencoder with our concrete selection layer, and performing both input reconstruction with the decoder reverse-pyramidal decoder $D$ as well as hazard prediction through the hazard prediction network $C$.

\subsection{Baselines}

\subsubsection{Maui} For this baseline \cite{ronen2019evaluation,uyar2021multi} we used the code made available by the original authors on github (\url{https://github.com/BIMSBbioinfo/maui}), and incorporated it into our testing pipeline as an integration method. We also adapted their cox-PH selection code to save the indexes which are selected as relevant for survival. We used our own Cox-regressor class on the significant factors, since it is equivalent to ones by Maui. The original Maui paper also used KMeans as a clusterer, but we limit our analyses to 2 subgroups, since this is a harder test for the model. Other small changes include changing their code to work with a different, more recent, version of Keras and Tensorflow. Note that, although we are using the Maui original code or adaptations of their code (where the original code does not store a model for later use), our pipeline is drastically different to Maui and more stringent, which might cause different performance to be reported here.

\subsubsection{Autoencoder + Cox-PH} Our AE baseline could be seen as a rough equivalent of \cite{lee2020incorporating}, and is the base on which our Supervised Autoencoder model was built. The model we've first attested following a similar approach \cite{chaudhary2018deep} uses an SGD optimiser instead of an Adam optimiser, and does so for a very small amount of training epochs, which in our initial testing proved to be equivalent to not training the algorithm at all, and might be seen as a form of random projections, like in \cite{bingham_random_2001}. Another approach used both omics-specific autoencoders (thus not performing cross-omics combinations in the fingerprints) and makes heavy use of boosting to improve the models joint performance, as well as still uses the SGD optimiser \cite{poirion2021deepprog}. Using code available on the model's github page (\url{https://github.com/lanagarmire/DeepProg}), we attested that the model provided similar outputs when not trained and when trained with the default number of epochs, most likely due to the use of SGD as an optimiser, which again makes the model be interpretable as a form of random projection  \cite{bingham_random_2001}. These models were not included in our final comparison due to the abovementioned methodological differences.

\subsubsection{PCA + Cox-PH} Our main baseline is the PCA baseline, which has been thoroughly used as a baseline in other papers, and was also used by the paper which first described the (unsupervised) Concrete Autoencoders \cite{balin2019concrete} as an upper-bound for the reconstruction loss of their CAE model. For the PCA baseline the first $d$ principal components are used as a drop-in equivalent for the $d$ fingerprint nodes in any of the autoencoder-based models.

\section{Results} \label{sec:results}

\subsection{Concordance Index Analysis}

\begin{figure*}
    \centering
    \includegraphics[width=.9\textwidth]{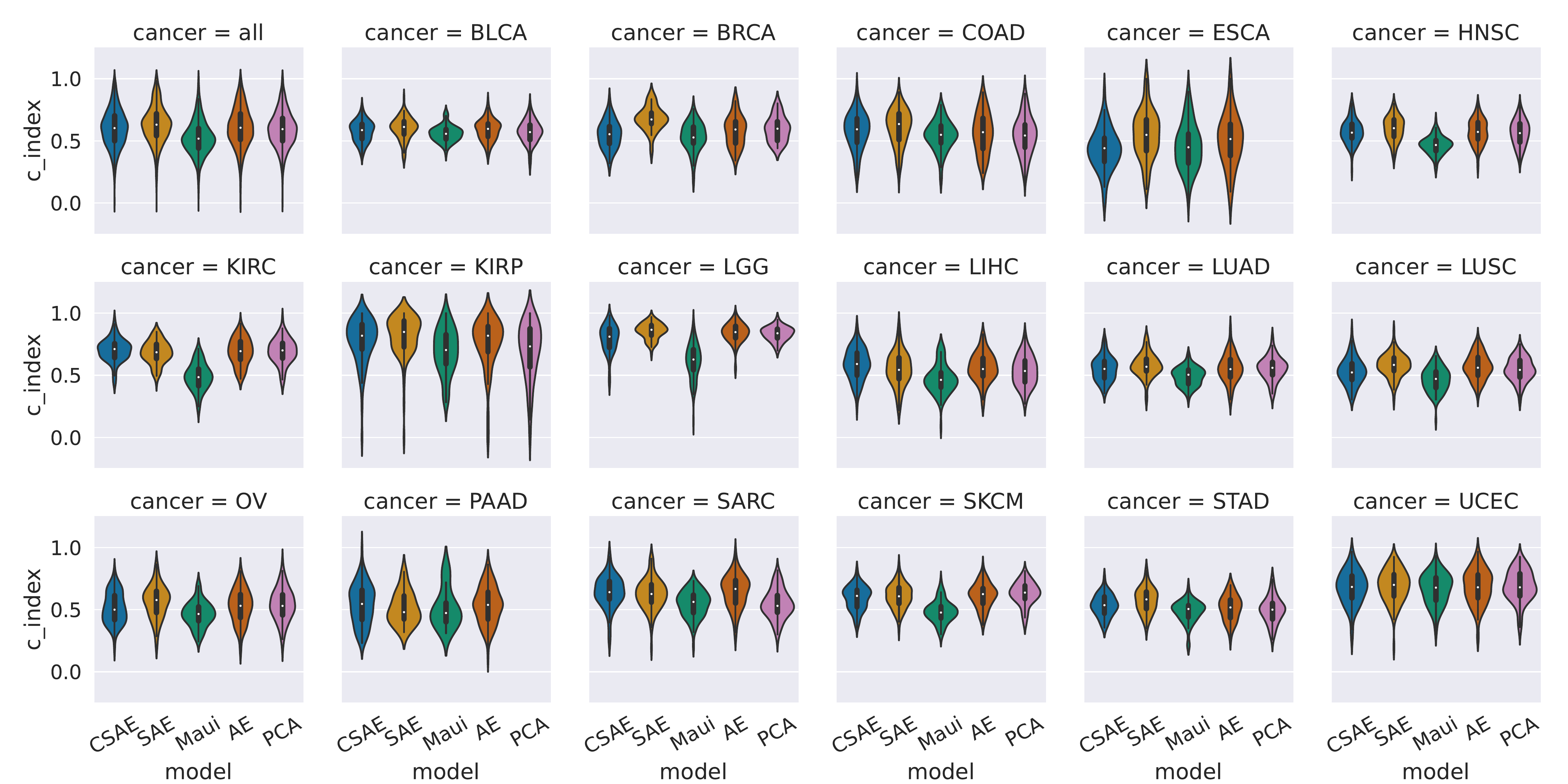}
    \caption{Violin plot of the Concordance Index (c\_index) cross-validation test performance of each method on each tested cancer Subtype as well as the Violin plot over all results. We can see that, even though our CSAE model relies on simpler feature selection instead of multiple nonlinear combinations, it manages to perform on-par with most other baselines.}
    \label{fig:c-index-comparison}
\end{figure*}

Using our stringent testing pipeline we evaluated all of the aforementioned models' capabilities to predict same-cancer out-of-sample instances through the cross-validation scheme. This is due to the fact that normalisation was calibrated using only training data, which is different, for example, to the methodology previously used for MAUI \cite{ronen2019evaluation}, where normalisation is done on all samples before validation, and the impact for this can be clearly seen on Figure~\ref{fig:c-index-comparison}, where we see that the concordance indexes reported in the original paper are higher than the ones we've encountered with this difference, something which was also noticed during our preliminary studies.

Also in Figure~\ref{fig:c-index-comparison}, we can see that our models either perform better than a PCA-based model or are not significantly different from the traditional PCA pipeline. In fact, Maui was the model which had the worst average rank, with 4.88
average rank, whereas the CSAE ranked equal to the PCA model at 2.94, only slightly worse than an AE model without Cox supervision with 2.53 average rank, with the cox-supervised autoencoder having 1.82 average rank.

With regards to overall Concordance-index, the previous ordering remains very much the same, with the SAE being the best overall with an average test score of 0.632 (all p-values statistically significant $<10^{-4}$ through an independent two-sided t-test), with the AE model having an average score of 0.610 (statistically significant difference $p<0.007$ to the Maui and PCA models), the CSAE with an average score of 0.603 (statistically significant difference $p<10^{-46}$ to the Maui model), and the Maui model having an average test C-index of 0.526.

These results tell us two things: the first one being that joint supervision on both the cox and reconstruction objectives improves out-of-sample performance in autoencoder based models. This was already somewhat attested previously \cite{tong_deep_2020,wissel_hierarchical_2022}, but none of these studies performed joint integration directly, using one autoencoder per omics layer \cite{tong_deep_2020}, which can be argued to not consist of integration at all, since the omics layers do not cross-contribute to the generated fingerprints, or having the generated fingerprints be de-factor integrated only through a two-step hierarchical step \cite{wissel_hierarchical_2022}. Here we attest that models that do perform this direct multi-omics integration, such as \cite{chaudhary2018deep,zhang2018deep,asada2020uncovering,lee2020incorporating,ronen2019evaluation,uyar2021multi} could greatly benefit from Cox Supervision as a joint optimisation loss.

\subsection{SKCM Analysis}

\begin{figure*}
    \centering
    \begin{subfigure}[b]{0.45\textwidth}
        \centering
        \includegraphics[width=\textwidth]{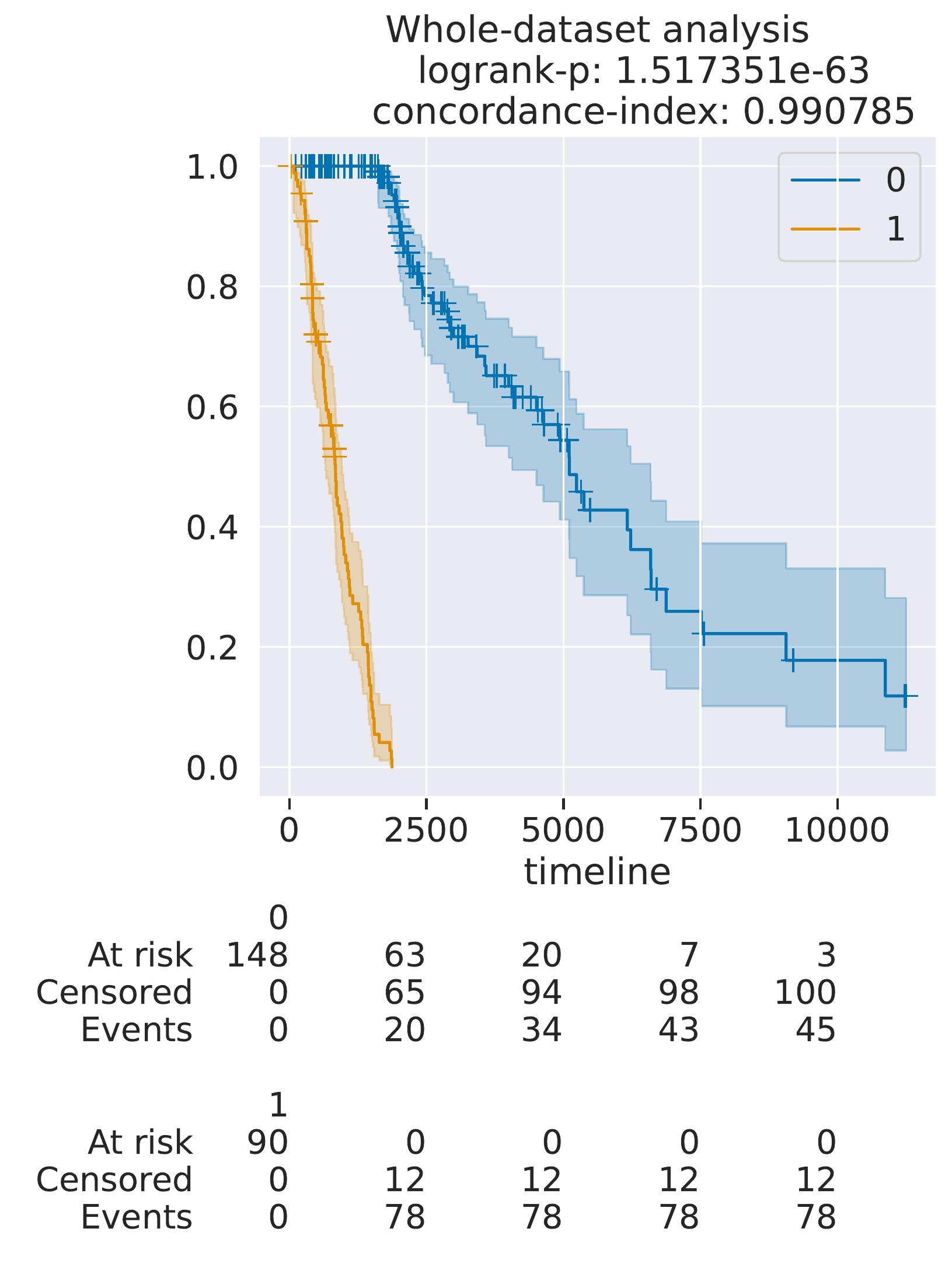}
        \caption{}
        \label{subfig:sae-whole-dataset-survival-separation}
    \end{subfigure}
    \begin{subfigure}[b]{0.45\textwidth}
        \centering
        \includegraphics[width=\textwidth]{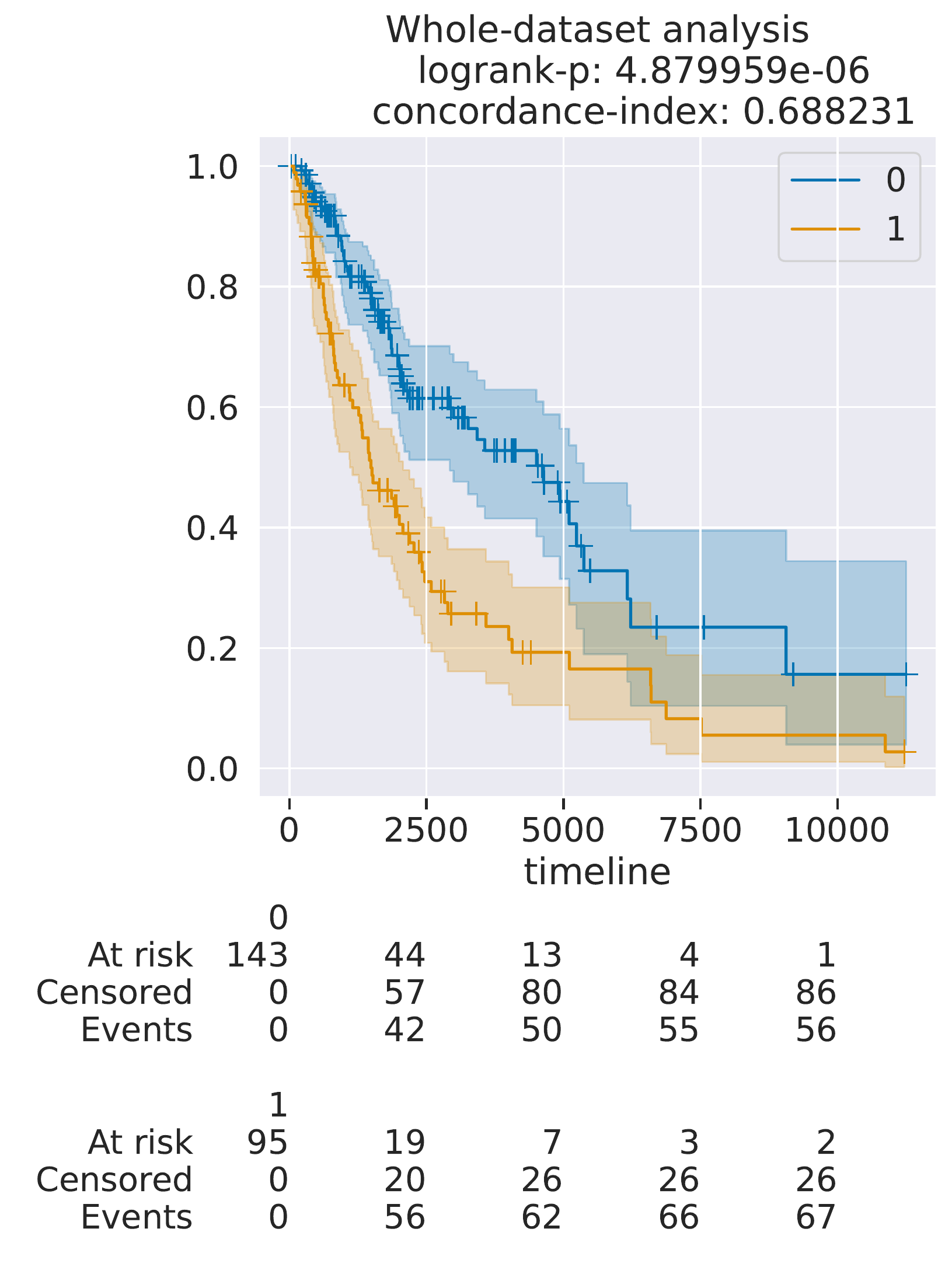}
        \caption{}
        \label{subfig:concretesae-whole-dataset-survival-separation}
    \end{subfigure}
    \caption{Survival curve separations for the SKCM dataset using our Supervised Autoencoder model \subref{subfig:sae-whole-dataset-survival-separation} and Concrete Supervised Autoencoder  \subref{subfig:concretesae-whole-dataset-survival-separation}}
    \label{fig:whole-dataset-survival-separation}
\end{figure*}

We can see that the Cox-Supervised Autoencoder fingerprints for the SKCM dataset provide a very clear separation in terms of survival outcomes. In Figure~\ref{subfig:sae-whole-dataset-survival-separation}, we see the 2-clustering results on the survival-relevant fingerprints for this model, which clearly separates a group of high-risk patients and a group of low-risk patients, despite it being 3rd place in terms of average C-index (only statistically significantly worse to PCA $p=0.02333$) provided a better separation on the Kaplain-Meier than what was previously reported with whole-dataset Kaplan-Meier plots in previous works (e.g., \cite{poirion2021deepprog}). The Concrete-SAE model, which was 4th place (but also only statistically significantly worse than PCA, $p=0.001685$), still manages an adequate survival stratification despite its ``fingerprints'' consisting of the original features, showing that it can still provide a good survival separation only on the basis of highly-interpretable feature selection (Figure~\ref{subfig:concretesae-whole-dataset-survival-separation}).

\begin{figure*}
    \centering
    \begin{subfigure}[b]{0.25\textwidth}
        \centering
        \includegraphics[width=\textwidth]{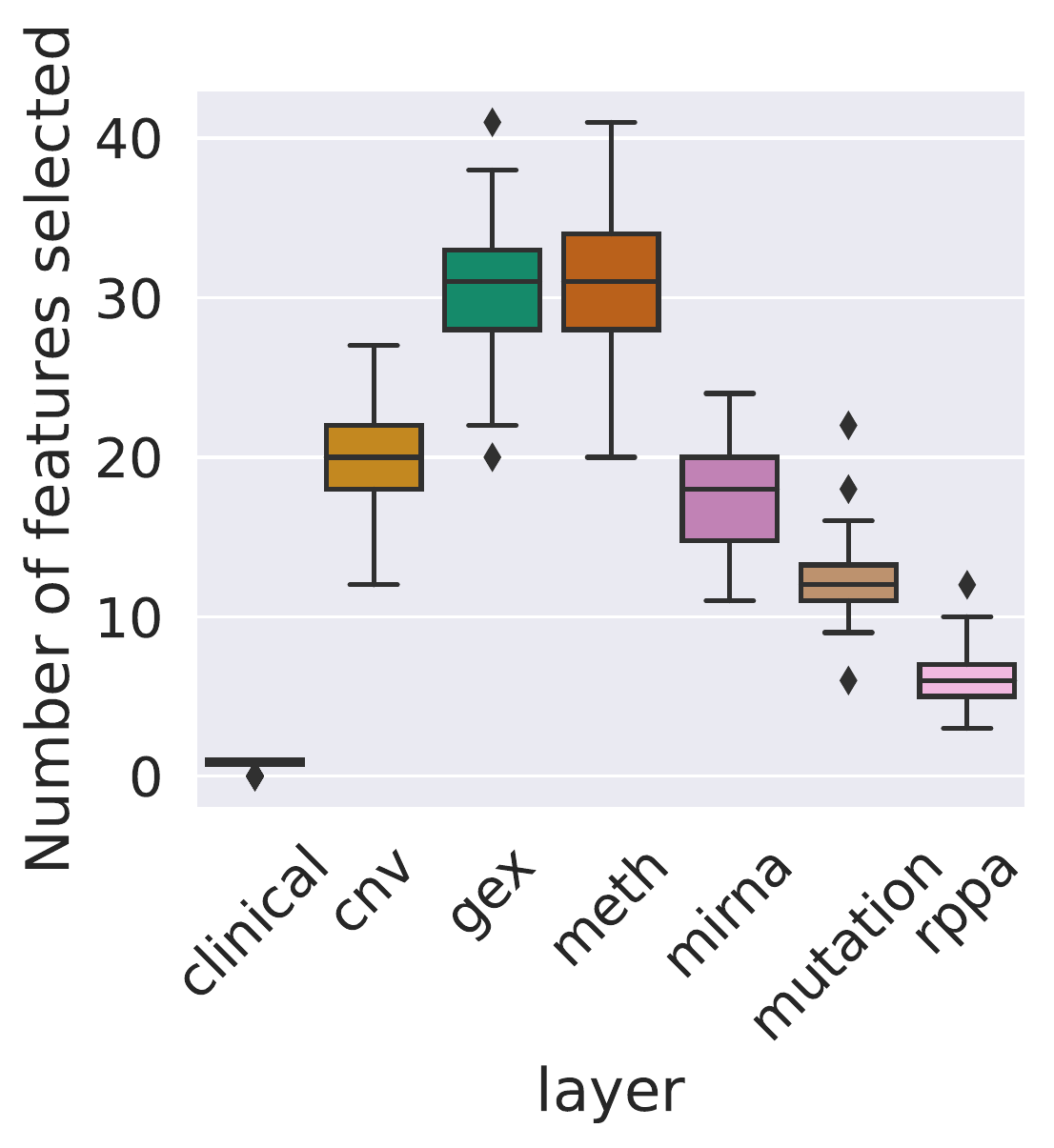}
        \caption{}
        \label{subfig:layer-distr-counts}
    \end{subfigure}
    \begin{subfigure}[b]{0.25\textwidth}
        \centering
        \includegraphics[width=\textwidth]{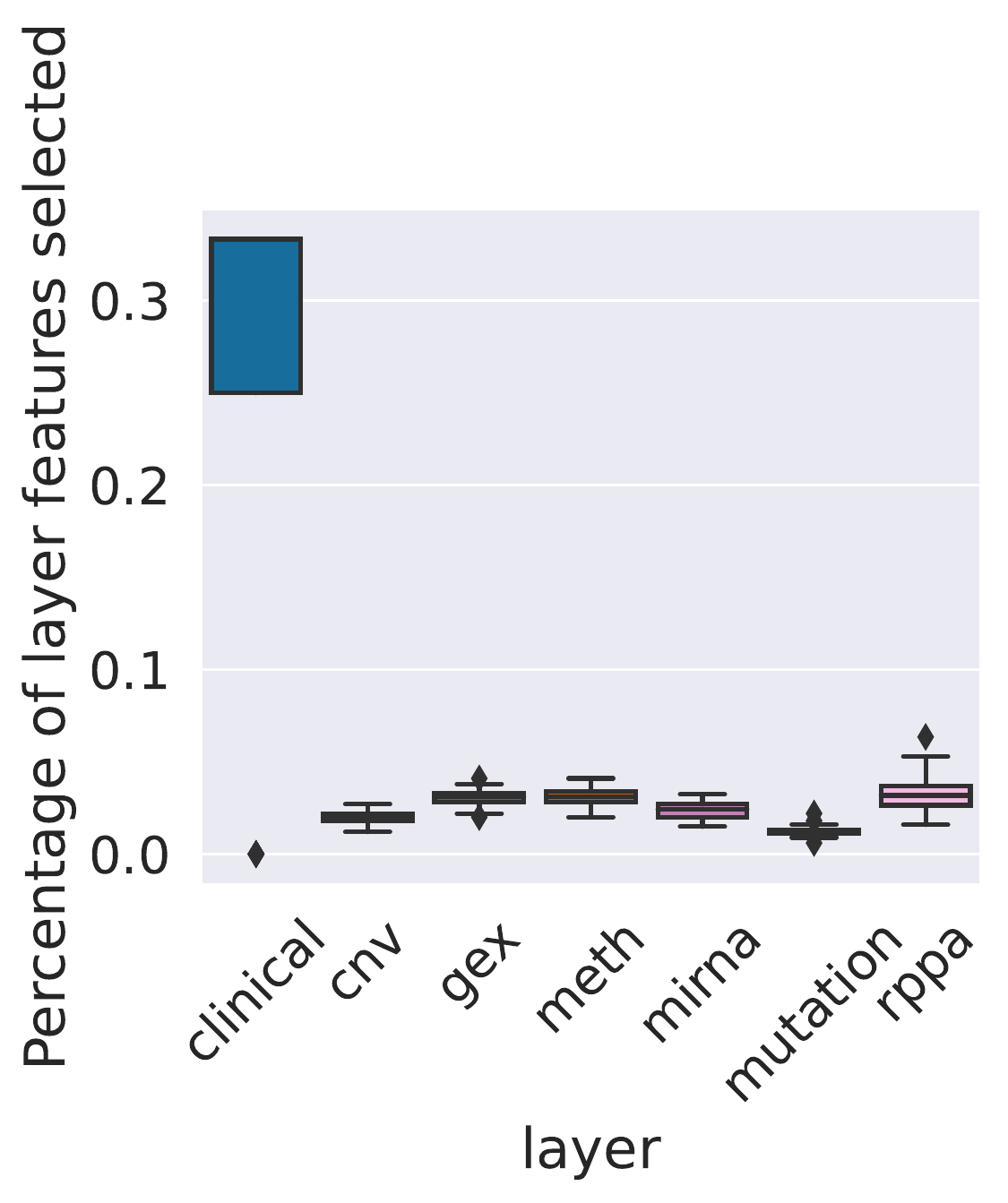}
        \caption{}
        \label{subfig:layer-distr-pct}
    \end{subfigure}
    \begin{subfigure}[b]{0.4\textwidth}
        \centering
        \includegraphics[width=\textwidth]{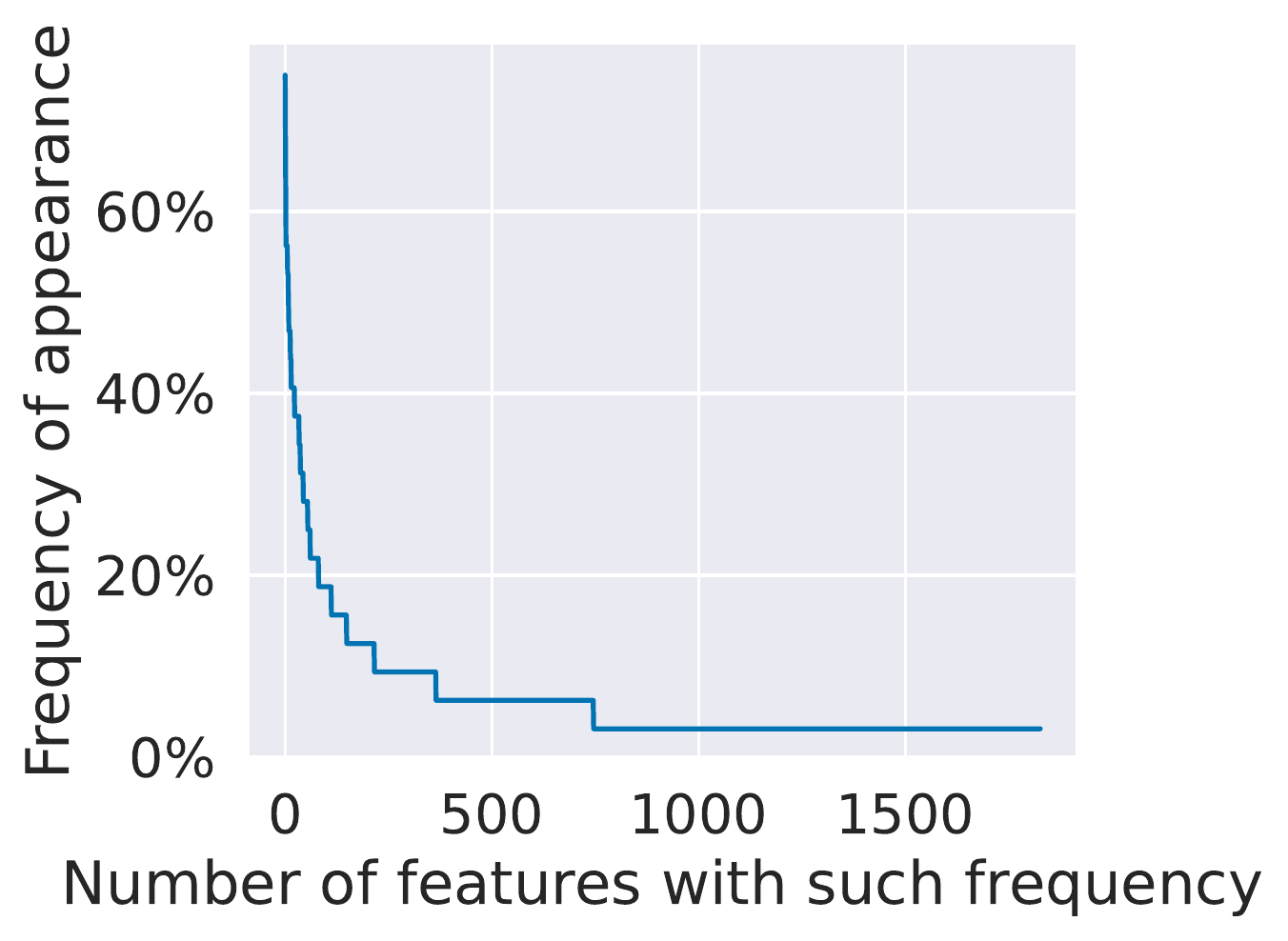}
        \caption{}
        \label{subfig:layer-distr-freqcurve}
    \end{subfigure}
    \caption{Layer feature selection distribution for the SKCM dataset. In \subref{subfig:layer-distr-counts} we can see how many features were selected by layer, with \subref{subfig:layer-distr-pct} showing the same information normalised by omics layer size, and \subref{subfig:layer-distr-freqcurve} showing the distribution of features that appear over multiple random initialisations of the algorithm.}
    \label{fig:layer-distr}
\end{figure*}

With regards to layer selections, we ran our models 32 times each and analysed the most important feature for each of their fingerprint features. In the case of the SAE model, possibly due to the fact that each fingerprint consists of a combination of combination of features, one feature consistently outranked the others and thus was used as the most important feature with regards to absolute neural path weight \cite{uyar2021multi}. The Concrete SAE model, however, since each fingerprint feature maps directly to an input feature, had a richer feature set selection, whose distributions can be seen in Figure~\ref{fig:layer-distr} and show a varied selection from multiple omics layers as well as a strong preference for a single Clinical factor which is very relevant for survival.

\section{Discussion}

In this paper we proposed two different models for Multi-Omics analysis: Our independently developed Cox-Supervised Autoencoder model, which is conceptually simpler than previously described models which also attempt at survival-based multi-omics analysis \cite{tong_deep_2020,wissel_hierarchical_2022}, proved to be very efficient, while providing true integration with regards to the sense adopted in previous work \cite{chaudhary2018deep,zhang2018deep,asada2020uncovering,lee2020incorporating,ronen2019evaluation,uyar2021multi}; our Concrete Cox-Supervised Autoencoder model, which performs Multi-Omics feature selection, instead, also proved to be a strong alternative for cases where interpretability is more favourable than expressive power, being more interpretable than, while being as powerful as, the PCA baseline, and not straying too far from its theoretical maximal baseline, our Cox-Supervised Autoencoder model.

Our proposed models, however, are not a one-size-fits-all solution to all survival-based multi-omics integration/feature selection challenges. Although one of our models ranked at least first or second with regards to survival separation on all but one dataset, the Concrete Supervised Autoencoder model is not expressive enough to capture cross-omics relationships due to its simple feature selection method, and our Supervised Autoencoder model might still be less expressive than its more complicated counterpart, the Hierarchical Supervised Autoencoder \cite{wissel_hierarchical_2022}, a comparison which we left for future work.

We believe that our Cox-Supervised Autoencoder model presented here provides a clear path forward, with a simple method for survival-based multi-omics integration, which can be further enriched with multitasking in its supervision, possibly also integrating drug responses, which can then lead to possible applications in drug discovery. Our Concrete Cox-Supervised Autoencoder model also makes use of recent advances \cite{maddison_concrete_2017,balin2019concrete} to provide an end-to-end differentiable feature selection, whose ramifications can range from finding specific sets of omics features that map to tasks other than survival, allowing us to leverage the power of differentiable programming techniques to discover new relationships in molecular assay datasets.

\section*{Acknowledgements} We would like to thank Dr Jonathan Cardoso-Silva for fruitful conversations throughout the development of this work, and João Nuno Beleza Oliveira Vidal Lourenço for designing the diagrams. We would also like to thank the developers and contributors of the lifelines (\url{https://github.com/CamDavidsonPilon/lifelines}) python package, which greatly facilitated the unification of the pipeline under a single python codebase. P.H.C.A. acknowledges that during his stay at KCL and A*STAR he's partly funded by King's College London and the A*STAR Research Attachment Programme (ARAP). The research was also supported by the National Institute for Health Research Biomedical Research Centre based at Guy's and St Thomas' NHS Foundation Trust and King's College London (IS-BRC-1215-20006). The authors are solely responsible for study design, data collection, analysis, decision to publish, and preparation of the manuscript. The views expressed are those of the author(s) and not necessarily those of the NHS, the NIHR or the Department of Health. This work used King's CREATE compute cluster for its experiments \cite{create_2022}. The results shown here are in whole or part based upon data generated by the TCGA Research Network: \url{https://www.cancer.gov/tcga}.

\bibliographystyle{plain}
\bibliography{dataintegration}

\end{document}